\documentclass[runningheads]{llncs}
\usepackage[T1]{fontenc}
\usepackage{graphicx}
\usepackage{CJKutf8} 
\begin{document}
\title{GPTs and Language Barrier: A Cross-Lingual \\Legal QA Examination}
%
%
\author{Ha-Thanh Nguyen${}^{1}$, Hiroaki Yamada${}^{2}$, Ken Satoh${}^{1}$}
\authorrunning{Nguyen et al.}
%
\institute{National Institute of Informatics \and
Tokyo Institute of Technology}
\maketitle              
\begin{abstract}
In this paper, we explore the application of Generative Pre-trained Transformers (GPTs) in cross-lingual legal Question-Answering (QA) systems using the COLIEE Task 4 dataset. In the COLIEE Task 4, given a statement and a set of related legal articles that serve as context, the objective is to determine whether the statement is legally valid, i.e., if it can be inferred from the provided contextual articles or not, which is also known as an entailment task. By benchmarking four different combinations of English and Japanese prompts and data, we provide valuable insights into GPTs' performance in multilingual legal QA scenarios, contributing to the development of more efficient and accurate cross-lingual QA solutions in the legal domain.

\end{abstract}
\section{Introduction}
The rapid increase in cross-border transactions and the globalization of legal systems have highlighted the importance of effective legal information retrieval across various languages. Legal Question-Answering (QA) systems assist professionals and individuals in locating relevant legal resources and extracting information to resolve disputes, comply with regulations, or make informed decisions. However, the vast amount of legal information available in diverse languages poses significant challenges to the development of efficient and accurate cross-lingual QA systems.

Current legal QA systems predominantly focus on mono-lingual information retrieval, which limits their capacity to address the needs of an increasingly multilingual and multicultural user base. Developing cross-lingual legal QA systems necessitates a deep understanding of not only the linguistic nuances but also the cultural and legal differences between the source and target languages. This requires advanced natural language processing techniques and robust models to accurately and effectively retrieve relevant information across languages.

In this study, we aim to investigate the potential of Generative Pre-trained Transformers (GPTs) in addressing the aforementioned challenges in cross-lingual legal QA systems. We utilize the COLIEE Task 4 dataset, which is derived from the Japanese Bar Exam and originally in Japanese, with the English version professionally translated by legal experts. The dataset comprises legal QA data related to entailment tasks in both languages. Task 4 involves determining the legal validity of a given statement based on a set of contextual legal articles. We evaluate the performance of GPTs under various settings, including different combinations of prompts and data across these languages, to gain insights into their effectiveness in cross-lingual legal QA scenarios.

Our research contributes to the broader understanding of GPTs' capabilities and limitations in cross-lingual legal QA scenarios, potentially paving the way for the development of more efficient and accurate multilingual legal QA systems. It also contributes to bridging the language gap in the legal domain, ultimately making legal information more accessible and comprehensible for users, regardless of their linguistic background.

\section{Related Work}

Research in the legal domain has presented numerous challenges for the Natural Language Processing (NLP) community, primarily due to the complexity of legal language and the need for contextual understanding in legal reasoning. As such, numerous studies have been conducted to improve the capacity of NLP models for legal tasks.

Legal textual entailment, a prominent area of research, has utilized the COLIEE Task 4 dataset \cite{Rabelo_2020,Rabelo_2022} as a crucial benchmark. Various methods have been employed to tackle this task, including traditional NLP techniques such as keyword-based matching \cite{kano2015keyword} and rule-based systems \cite{rosa2021yes} for information extraction and entailment identification from legal documents. However, these approaches may suffer from rigidity in methodology and limited flexibility in responding to subtle changes in language patterns and context.

With the advent of word embeddings like Word2Vec \cite{mikolov2013efficient} and transformers, such as BERT \cite{devlin2018bert}, RoBERTa \cite{liu2019roberta}, and ALBERT \cite{lan2019albert}, NLP made significant advancements, leading to improved performance across various legal tasks. As a result, there have been studies on different tasks like legal lawfulness classification \cite{nguyen2019deep}, information extraction \cite{zin2023improving}, question-answering \cite{kien2020answering}, and multi-task learning \cite{vuong2023nowj} that leverage these techniques.

Few-shot learning and few-relational learning strategies have also been introduced in legal entailment tasks \cite{bilgin2023amhr,rosa2022billions}. With these approaches, models can learn from a limited amount of labeled data to generalize for unseen cases, which is invaluable given the scarcity of high-quality, annotated legal documents.

Recent GPT models have shown impressive performance in various NLP tasks, including statutory reasoning tasks \cite{blair2023can,bilgin2023amhr}, further demonstrating their applicability in the legal domain. GPT-4, specifically, has delivered remarkable results on the Uniform Bar Examination (UBE) \cite{katz2023gpt}, showcasing the potential of GPT models in various legal reasoning tasks \cite{nguyen2023logilaw,nguyen2023lawgiba}.

Building upon these advancements, our study aims to evaluate the performance of GPT-3.5 and GPT-4 on the legal textual entailment task using the COLIEE dataset, focusing on both monolingual and cross-lingual aspects. We analyze the models' performance across different dataset versions, ranging from Heisei 18 (2006) to Reiwa 3 (2021), to gain a better understanding of the potential challenges and capabilities of these state-of-the-art models in the complex legal domain. This in-depth analysis, along with the exploration of cross-lingual and global law aspects, provides valuable insights into the prospects of employing GPT models in legal QA systems for various languages and jurisdictions.

\section{Experiment Design}

In this section, we detail the experimental settings used to evaluate the performance of GPT-4 and GPT-3.5 models in monolingual and cross-lingual scenarios. We utilize the COLIEE Task 4 dataset, which provides legal questions and answers in both English and Japanese.

\subsection{Data Analysis}

The dataset comprises data from different years: H29, H30, R01, R02, and R03, with a varying number of questions each year, as displayed in Table \ref{tab:question_count}.

\begin{table}[t]
\centering
\begin{tabular}{lr}
\hline
\textbf{Year} & \textbf{Question Count} \\
\hline
H29 & 58 \\
H30 & 70 \\
R01 & 111 \\
R02 & 81 \\
R03 & 109 \\
\hline
\end{tabular}
\caption{Number of questions for each year in the dataset.}
\label{tab:question_count}
\end{table}

Further, we present the length statistics of the dataset in English and Japanese languages in Tables \ref{tab:english_length_stats} and \ref{tab:japanese_length_stats}, respectively. The statistics provide insights into the dataset's characteristics, giving us a better understanding of the textual complexity and sizing in each language.

\begin{table}[ht]
\centering
\begin{tabular}{ccccccc}
\hline
\textbf{Year} & \multicolumn{3}{c}{\textbf{Context Length}} & \multicolumn{3}{c}{\textbf{Question Length}} \\
\cline{2-4}
\cline{5-7}
 & \textbf{Min} & \textbf{Avg} & \textbf{Max} & \textbf{Min} & \textbf{Avg} & \textbf{Max} \\ \hline
H29 & 118 & 655.31 & 2505 & 60 & 273.86 & 601 \\
H30 & 113 & 525.10 & 1682 & 40 & 207.24 & 847 \\
R01 & 109 & 536.04 & 2604 & 44 & 200.49 & 501 \\
R02 & 99 & 541.62 & 1880 & 60 & 215.78 & 379 \\
R03 & 68 & 702.99 & 4048 & 66 & 234.62 & 558 \\ \hline
\end{tabular}
\caption{English dataset character length statistics.}
\label{tab:english_length_stats}
\end{table}

\begin{table}[t]
\centering
\begin{tabular}{ccccccc}
\hline
\textbf{Year} & \multicolumn{3}{c}{\textbf{Context Length}} & \multicolumn{3}{c}{\textbf{Question Length}} \\
\cline{2-4}
\cline{5-7}
 & \textbf{Min} & \textbf{Avg} & \textbf{Max} & \textbf{Min} & \textbf{Avg} & \textbf{Max} \\ \hline
H29 & 37 & 165.74 & 566 & 28 & 87.16 & 198 \\
H30 & 18 & 110.46 & 409 & 19 & 72.84 & 285 \\
R01 & 39 & 168.39 & 728 & 15 & 73.20 & 223 \\
R02 & 37 & 167.57 & 536 & 21 & 88.56 & 219 \\
R03 & 29 & 213.28 & 1145 & 29 & 77.88 & 173 \\ \hline
\end{tabular}
\caption{Japanese dataset character length statistics.}
\label{tab:japanese_length_stats}
\end{table}

The data indicate ranges of average lengths for context in English and Japanese as follows:
\begin{itemize}
\item English context length varies from 525 characters (H30) to 703 characters (R03).
\item Japanese context length ranges from 110 characters (H30) to 213 characters (R03).
\end{itemize}

English question average lengths fluctuate between 200 characters (R01) and 273 characters (H29), while Japanese question lengths range from 72 characters (H30) to 88 characters (R02). Analyzing these variations in average lengths of both context and question in different years is important since it provides insights into the dataset's complexity and potential text processing challenges.

Looking at Tables \ref{tab:english_length_stats} and \ref{tab:japanese_length_stats}, we can see that the maximum length of context is 4,048 characters and that of questions is 847 characters, which is considerably less than the token limit of GPT-4 (0314) at 8,192 tokens and GPT-3.5 (Turbo) at 4,191 tokens. Additionally, it should be noted that one token may contain one or more characters\footnote{https://github.com/openai/tiktoken}.

\subsection{Monolingual and Cross-lingual Prompting}
We formatted the input prompt as follows for monolingual prompting:

\noindent \textbf{Prompt in English:}
\begin{quote}
\texttt{\{context\}\\
Question: \{question\}\\
Answer (Y or N), no explain.}
\end{quote}

\begin{CJK*}{UTF8}{min}
\noindent \textbf{Prompt in Japanese:}
\begin{quote}
\texttt{\{context\}\\
質問: \{question\}\\
回答 (Y または N)、説明は不要。}
\end{quote}
\end{CJK*}

Here, \texttt{context} represents the relevant articles, and \texttt{question} corresponds to a given question in the dataset. This prompt format ensures consistency in the input and allows the models to focus solely on providing a binary answer—either ``Y'' (YES) or ``N'' (NO).

In our experiments, we explore different combinations of context and question languages, yielding four distinct settings: English context and English question (EN-EN), Japanese context and Japanese question (JA-JA), and two cross-lingual settings: English context with Japanese question (EN-JA) and Japanese context with English question (JA-EN).

By evaluating the models in both monolingual (EN-EN, JA-JA) and cross-lingual (EN-JA, JA-EN) settings, we aim to assess how well GPT-3.5 (ChatGPT) and GPT-4 can handle linguistic nuances, cultural differences, and mappings between languages in the legal domain. This information will provide valuable insights into their potential for deployment in multilingual legal QA systems.

With a focus on GPT-4 and GPT-3.5 models, our experiments aim to examine their performance under these diverse conditions, providing valuable insights into their potential for deployment in multilingual legal QA systems.

\section{Experimental Results}

\begin{table*}
\centering
\begin{tabular}{lccccc}
\hline
 & \multicolumn{5}{c}{\textbf{Years}} \\
\cline{2-6}
\textbf{Model}                    & \textbf{H29} & \textbf{H30} & \textbf{R01} & \textbf{R02} & \textbf{R03} \\ \hline
GPT-4 JA-EN (Cross-Lingual)    & 0.7241  & 0.6857  & 0.7658  & 0.7901  & 0.8349  \\
GPT-4 EN-JA (Cross-Lingual)    & 0.6724  & 0.7571  & 0.8198  & 0.7901  & 0.7798  \\
GPT-3.5 JA-EN (Cross-Lingual)  & 0.6034  & 0.6571  & 0.5946  & 0.6420  & 0.6697  \\
GPT-3.5 EN-JA (Cross-Lingual)  & 0.5690  & 0.7143  & 0.6937  & 0.6049  & 0.6514  \\ \hline
GPT-3.5 EN (Monolingual)     & 0.6207  & 0.5857  & 0.6486  & 0.6914  & 0.7156  \\
GPT-4 EN (Monolingual)       & 0.7414  & 0.7429  & 0.8108  & 0.8148  & 0.8440  \\
GPT-3.5 JA (Monolingual)      & 0.6207  & 0.6714  & 0.5676  & 0.5926  & 0.6514  \\
GPT-4 JA (Monolingual)        & 0.7586  & 0.7857  & 0.7838  & 0.8642  & 0.8807  \\ \hline
\end{tabular}
\caption{Performance Comparison in Monolingual and Cross-Lingual Settings by GPT-3.5 and GPT-4.}
\label{tab:model_perf_comparison}
\end{table*}

\begin{figure*}
    \centering
    \includegraphics[width=.8\textwidth]{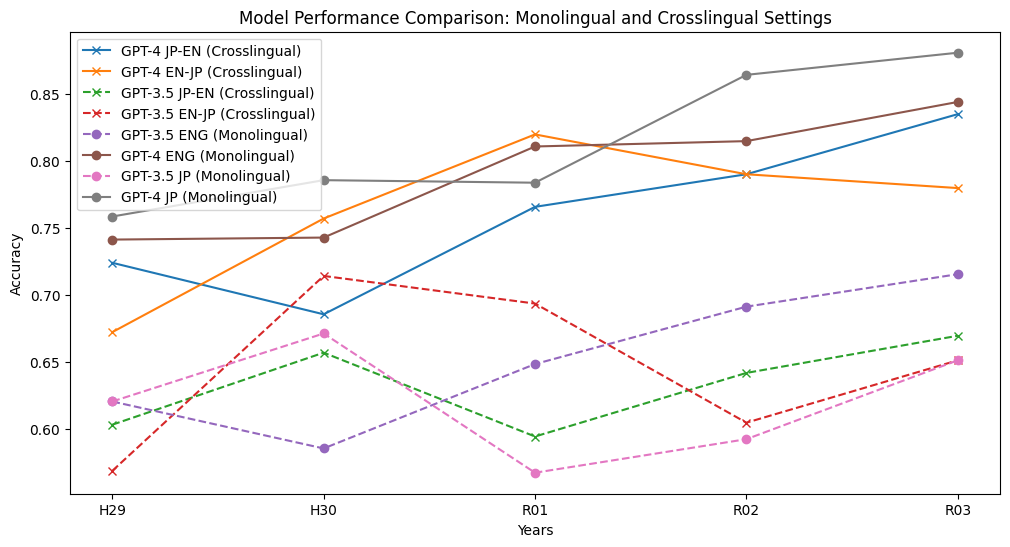}
    \caption{Performance Comparison Data Visualization.}
    \label{fig:model_perf_comparison_plot}
\end{figure*}

In Table~\ref{tab:model_perf_comparison} and Figure~\ref{fig:model_perf_comparison_plot}, we compare the performance of GPT-4 (gpt-4-0314) and GPT-3.5 (gpt-3.5-turbo) in monolingual and cross-lingual settings. The temperature is set at 1, Top P is set at 1, the frequency penalty is 0, and the presence penalty is 0 for both models. While it is important to note that the years H29 to R03 are independent and not related, we can still observe certain trends and patterns in the results.

It is evident that the GPT-4 model consistently outperforms the GPT-3.5 model across all independent yearly instances in both monolingual and cross-lingual settings. This superior performance could be attributed to the advancements in the model architecture and higher capacity, allowing GPT-4 to capture the complexities of the languages and tasks more effectively.

Furthermore, monolingual settings generally yield higher accuracy scores than cross-lingual settings for both models. This can be attributed to the fact that monolingual settings tend to be more straightforward, not requiring translation within the same language. On the other hand, cross-lingual settings involve handling linguistic nuances, cultural differences, and complex mappings between languages, making the task more challenging for even advanced models like GPT-4 and GPT-3.5.

The observation that Japanese monolingual performance is better than English monolingual performance can also be explained by the nature of the original data being in Japanese. Given that the source data is in Japanese, it is reasonable to infer that the models may have been more effective in understanding and processing the original data in Japanese text. The derived English translations primarily serve as reference material and might not capture the fine-grained nuances as effectively as the original Japanese text, contributing to the observed difference in performance. This observation is also consistent with the results of COLIEE competitions in recent years, where teams using Japanese as their primary language tend to outperform those using English \cite{Rabelo_2020,Rabelo_2022}. 

This finding aligns with the, earlier mentioned, superior performance of monolingual settings compared to cross-lingual ones and highlights the importance of having high-quality translated material and a deep understanding of the languages involved when working with cross-lingual tasks. It also emphasizes the challenges faced by models like GPT-4 and GPT-3.5 in adapting to and dealing with different natural languages and contextual information, further motivating research and development efforts to improve their performance for cross-lingual tasks.

The experimental results demonstrate the prowess of the GPT-4 model over its predecessor, GPT-3.5, in both monolingual and cross-lingual settings. As advancements in model architecture and training techniques continue to be made, we can expect further performance improvements. However, achieving human-level performance in cross-lingual settings remains a challenge and may demand more significant advancements in model design, access to high-quality datasets, and a deeper understanding of the complexities involved in cross-lingual tasks.

\section{Conclusions}
In this study, we evaluated GPT-3.5 and GPT-4 models in cross-lingual legal Question-Answering using the COLIEE Task 4 dataset. Our findings showed the superior performance of GPT-4 over GPT-3.5 and highlighted the challenges faced by GPT models in handling linguistic nuances
in cross-lingual settings. These insights contribute to the development of more accurate cross-lingual QA systems in the legal domain and emphasize the importance of high-quality translation and deep understanding of linguistic complexities. Future research should focus on improving GPT models' cross-lingual capabilities and incorporating domain-specific knowledge tailored to the legal field.

\section*{Acknowledgements}
This work was supported by the AIP challenge funding related with JST, AIP Trilateral AI Research, Grant Number JPMJCR20G4.

\bibliographystyle{splncs04}
\bibliography{ref}
\end{document}